\documentclass{article}

\usepackage[final,nonatbib]{neurips_2021}

\usepackage[style=numeric,sorting=none]{biblatex}
\usepackage[utf8]{inputenc} % allow utf-8 input
\usepackage[T1]{fontenc}    % use 8-bit T1 fonts
\usepackage{hyperref}       % hyperlinks
\usepackage{url}            % simple URL typesetting
\usepackage{booktabs}       % professional-quality tables
\usepackage{amsfonts}       % blackboard math symbols
\usepackage{nicefrac}       % compact symbols for 1/2, etc.
\usepackage{microtype}      % microtypography
\usepackage{xcolor}         % colors
\usepackage{amsmath,amsfonts,bm}

\usepackage{xcolor}

\def\eqref#1{Eq.~(\ref{#1})}
\def\1{\bm{1}}

\def\vp{{\bm{p}}}
\def\vq{{\bm{q}}}

\DeclareMathAlphabet{\mathsfit}{\encodingdefault}{\sfdefault}{m}{sl}
\SetMathAlphabet{\mathsfit}{bold}{\encodingdefault}{\sfdefault}{bx}{n}

\newcommand{\bp}{\bs{\phi}}

\newcommand{\be}{\begin{equation}}
\newcommand{\ee}{\end{equation}}
\newcommand{\bc}{\begin{center}}
\newcommand{\ec}{\end{center}}
\newcommand{\bd}{\begin{description}}
\newcommand{\ed}{\end{description}}
\newcommand{\bi}{\begin{itemize}}
\newcommand{\ei}{\end{itemize}}
\newcommand{\pa}{\partial}
\newcommand{\bs}{\boldsymbol}

\def\RR{ \mathbb R}

\newcommand{\bmat}{\begin{pmatrix}}
\newcommand{\emat}{\end{pmatrix}}
\newcommand{\bsmat}{\left(\begin{smallmatrix}}
\newcommand{\esmat}{\end{smallmatrix}\right)}
\newcommand{\bes}{\begin{equation}\begin{split}}
\newcommand{\ees}{\end{split}\end{equation}}

\newcommand{\Norm}[1]{\mathinner{\left\Vert#1\right\Vert}}%
\let\vec\relax%
\newcommand{\vec}[1]{\ensuremath{\boldsymbol{#1}}}%
\def\vecdeflatinlower#1{\expandafter\def\csname v#1\endcsname{\vec{#1}}}%
\def\vecdeflatinlowerall#1{\ifx#1\vecdeflatinlowerall\else\vecdeflatinlower{#1}\expandafter\vecdeflatinlowerall\fi}%
\vecdeflatinlowerall abcdefghijklmnopqrstuvwxyz\vecdeflatinlowerall%
\def\vecdeflatinupper#1{\expandafter\def\csname v#1\endcsname{\vec{#1}}}%
\def\vecdeflatinupperall#1{\ifx#1\vecdeflatinupperall\else\vecdeflatinupper{#1}\expandafter\vecdeflatinupperall\fi}%
\vecdeflatinupperall ABCDEFGHIJKLMNOPQRSTUVWXYZ\vecdeflatinupperall%
\def\vecdefgreeklower#1{\expandafter\def\csname v#1\endcsname{\vec{\csname #1\endcsname}}}%
\def\vecdefgreeklowerall#1{\ifx#1\vecdefgreeklowerall\else\vecdefgreeklower{#1}\expandafter\vecdefgreeklowerall\fi}%
\vecdefgreeklowerall {alpha}{beta}{gamma}{delta}{epsilon}{varepsilon}{zeta}{eta}{theta}{vartheta}{iota}{kappa}{lambda}{mu}{nu}{pi}{varpi}%
{xi}{rho}{varrho}{sigma}{varsigma}{tau}{upsilon}{phi}{varphi}{chi}{psi}{omega}\vecdefgreeklowerall%
\def\vecdefgreekupper#1{\expandafter\def\csname v#1\endcsname{\vec{\csname #1\endcsname}}}%
\def\vecdefgreekupperall#1{\ifx#1\vecdefgreekupperall\else\vecdefgreekupper{#1}\expandafter\vecdefgreekupperall\fi}%
\vecdefgreekupperall {Gamma}{Delta}{Theta}{Lambda}{Xi}{Pi}{Sigma}{Upsilon}{Phi}{Psi}{Omega}\vecdefgreekupperall%
\newcommand{\pdiff}[3][]{\frac{\partial^{#1}\,\!#2}{\partial\,\!#3^{#1}}}%
\def\inv#1{\frac{1}{#1}}	 												% inverse 1/x 

\title{Physics-enhanced Neural Networks in the Small Data Regime}

\author{%
  Jonas Eichelsdörfer\thanks{equal contribution} \,, \, Sebastian Kaltenbach\footnotemark[1] \,, \, Phaedon-Stelios Koutsourelakis \\
Professorship of Continuum Mechanics\\
Technical University of Munich\\
Munich, Germany\\
\texttt{\{jonas.eichelsdoerfer,sebastian.kaltenbach,p.s.koutsourelakis\}@tum.de} \\
}

\begin{document}

\maketitle

\begin{abstract}
 Identifying the dynamics of physical systems requires a machine learning model that can assimilate observational data, but also incorporate the laws of physics. Neural Networks based on physical principles such as the Hamiltonian or Lagrangian NNs have recently shown promising results in generating extrapolative predictions and accurately representing the system's dynamics. We show that by additionally considering the actual energy level as a regularization term during training and thus using physical information as inductive bias, the results can be further improved. Especially in the case where  only small amounts of data are available, these improvements  can significantly enhance the predictive capability.
 We apply the proposed regularization term to a Hamiltonian Neural Network (HNN) and Constrained Hamiltonian Neural Network (CHHN) for a single and double pendulum, generate predictions under unseen initial conditions and report significant  gains in predictive accuracy.
\end{abstract}

\section{Introduction}
The present paper is concerned with the discovery of the dynamics of physical systems in the Small Data regime. Modern machine learning techniques without any inductive bias often fail in this task because %the cost of data acquisition is prohibitive or 
the networks do not generalize well and lack robustness. To overcome these problems, we therefore combine latest advances in physics-enhanced Neural Networks based on dynamic invariants such as the Hamiltonian \cite{Greydanus.2019,toth2019hamiltonian,Finzi.2020} or Lagrangian \cite{lutter2019deep,Cranmer.2020} NNs with additional inductive bias to gain better predictive capabilities and to reduce the requisite training data. We propose a novel regularization term that can be readily added to the training  loss function of most machine learning frameworks and  show that it leads  to predictive performance and data efficiency that outperforms  the original framework for all tested examples and under initial conditions not contained in the training data.

This work shares similarities with approaches that use physical laws as regularization terms or  augment the loss function as in \cite{raissi2019physics,lusch2018deep,zhu2019physics,kaltenbach2020incorporating}. Another line of work closely related  to ours pertains to the use of inductive bias with regards to the long-term stability  \cite{kaltenbach_2021} or embedded symmetries \cite{walters_2021} of physical systems. Moreover, Graph Neural Networks \cite{Scarselli.2008, Battaglia.2018} can be tailored to physical settings by  e.g. combining them with the aforementioned Hamiltonian \cite{SanchezGonzalez.2019}.

\section{Physics-enhanced Neural Networks}
Domain knowledge originating from physics  can be used in various ways to enhance the performance of neural networks. This section introduces physics-enhanced neural networks based on dynamic invariants and a novel regularization term based on the energy level/difference of the training data.

\subsection{Hamiltonian Neural Networks}
The basic idea of  HNNs is to incorporate the principle of Hamiltonian mechanics as inductive bias. Instead of approximating the governing equations for each degree of freedom, the Hamiltonian equations are invoked, i.e. $\dot{\vq}=\frac{\pa H}{\pa \vp}, ~\dot{\vp}=\frac{\pa H}{\pa \vq}$, and solely % of a physical system directly,
a differentiable representation of the system's Hamiltonian $H$ is learned. We note that for isolated systems,  the Hamiltonian is written  in terms of generalized coordinates $\bs{q}$ and momenta $\bs{p}$, and  it expresses the system's total energy which is conserved during the dynamic evolution.\\
Greydanus \cite{Greydanus.2019} proposed to learn the Hamiltonian   on the basis of $N$ observations  of the system's state variables $\bs{z}^{(i)}=(\bs{q}^{(i)},\bs{p}^{(i)})$ as  well as their time derivatives $\dot{\bs{z}}^{(i)}=(\dot{\bs{q}}^{(i)},\dot{\bs{p}}^{(i)}), i=1,\ldots,N$ %from the system's state and associated time derivative data pairs $((\vq, \vp),(\vec{\dot{q}}, \vec{\dot{p}}))_i$
with the following loss function $\mathcal{L}$:
\begin{equation}
	\mathcal{L}(\bp) =\inv{N} \sum_{i=1}^{N} \Norm{\pdiff{H_{\bp}(\bs{z}^{(i)})}{\vp} - \dot{\vq}^{(i)}}^2 + \Norm{\pdiff{H_{\bp}(\bs{z}^{(i)})}{\vq} + \dot{\vp}^{(i)} }^2
\end{equation}
where $\bp$ are the tunabe parameters (e.g. NN weights/biases).
Since the Hamiltonian dynamics automatically respect the conservation of energy, a significant improvement in long term simulation accuracy can be achieved. As the loss function is not based on the absolute value of $H$ value but its gradient, the learned Hamiltonian may differ from the system's total energy by a constant factor.\\
We note that the accuracy in the learned $H$ must be extremely high in order to provide good estimates of its derivatives which are needed for predictive purposes and which becomes increasingly challenging as the dimension of $\bs{z}$ increases.

\subsection{Constrained HNNs}
Another promising approach to represent the Hamiltonian of a physical system is the constrained Hamiltonian method. Finzi et al. \cite{Finzi.2020} recognized that frequently the structure of the Hamiltonian is significantly simpler if expressed in Cartesian rather than generalized coordinates and achieved better and more data-efficient results than the HNN method in multibody dynamics. The algorithm proposed employs  a mapping from  the Cartesian coordinates onto the constraint manifold described by the generalized coordinates. The downside of the method is that in order to use this the mapping, a linear system of equations must be solved at every inference step, both during the network training and prediction.

\subsection{Lagrangian Neural Networks}
To overcome the restriction of employing generalized  coordinates in the Hamiltonian formalism, Cranmer et al.\cite{Cranmer.2020} proposed  Lagrangian Neural Networks (LNNs). Based on the Lagrangian the acceleration of any degree of freedom   can readily be calculated and a loss function can be stated in terms of the discrepancy between the true and the model-predicted acceleration. %, based on the learned Lagrangian. 
This method therefore is more generally applicable than the Hamiltonian formalism, but does involve a matrix inversion during each training step \cite{Cranmer.2020}. 

\subsection{Proposed regularization}\label{sec:Hamiltonian_Neural_Network}
In this work, the loss function of physics-enhanced neural networks such as HNN and CHNN is extended by an additional, physics-informed, regularization term which  penalizes  the difference between the learned Hamiltonian $H_{\bp}(\bs{z})$ and target values of the system's total energy level $\hat{H}$, thereby setting the level of the predicted Hamiltonian. It is sufficient to compute $\hat{H}$ once for every initial condition contained in the training data. The proposed regularization term restricts the inferred solution to the numerical value of the total energy data. The augmented loss function is given by 
\begin{equation}\label{eq:HNNloss}
	\bar{\mathcal{L}}({\bp}) = \inv{N} \sum_{i=1}^{N} \Norm{\pdiff{H_{\bp}(\bs{z}^{(i)})}{\vp} - \dot{\vq}^{(i)}}^2 + \Norm{\pdiff{H_{\bp}(\bs{z}^{(i)})}{\vq} + \dot{\vp}^{(i)} }^2+ \lambda_H (H_{\bp}(\bs{z}^{(i)}) - \hat{H}_i)^2 \enspace,
\end{equation} 

The total energy of the system must therefore be collected/computed as additional training data. We note that the zero level of this total energy can be set arbitrarily. As during training and for predictions only the gradients of the Hamiltonian are relevant, the actual zero level does not matter. From a physical point of view, we are passing information regarding the energy level difference between the different training data points to the algorithm. This regularization becomes increasingly important in the small-data regime. For real world problems, the generation of large amounts of training data is often prohibitively expensive. It is thus crucial to improve the robustness of the algorithms under limited training data. If data on the total energy of a system is available, it can thus be leveraged to improve the long term accuracy and generalization capability of the learned Hamiltonian dynamics as we demonstrated in the sequel.\\
The proposed regularization involves an additional hyperparameter $\lambda_H$. The optimal value of $\lambda_H$ is dependent on the problem at hand and has to be found by cross-validation.\\

\section{Numerical Illustrations}
The proposed regularization term is applied to HNNs and CHNNs for a single and a double pendulum. All numerical schemes discussed in this work are implemented with the help of the Jax framework \cite{JamesBradbury.2018}, a library for high-performance machine learning research based on the well-known Autograd \cite{Maclaurin.2015} and TensorFlow \cite{MartnAbadi.2015} projects. Training data was  generated by numerical integration of \eqref{eq:singlePendulumDynamics}, \eqref{eq:2pd}  with a $4^{th}$-order Runge-Kutta scheme and a time-step of $0.1s$. The results of a black-box MLP model \cite{Greydanus.2019,Finzi.2020} serve as baseline in Tables \ref{tab:singlePendulum_fullData_deltaH}, \ref{tab:doublePendulum_fullData_deltaH}. The code can be found at \url{https://github.com/pkmtum/Physics-enhanced_NN_SmallData}.

\subsection{Single Pendulum}
The first example considered is the single-degree-of-freedom pendulum which involves a point mass $m$, suspended from its pivot with a massless rod of length $l$, is swinging frictionlessly.
%under the influence  of gravity. The mass is suspended from its pivot with a massless rod.
When displaced, gravity causes the pendulum to oscillate periodically about the equilibrium position. The angle $\theta$ between the pendulum rod and the vertical axis is chosen as the sole generalized coordinate. The equation of motion reads
\begin{equation}\label{eq:singlePendulumDynamics}
	\ddot{\theta} + \frac{g}{l} \sin \theta = 0 .
\end{equation}  
The mass $m$ and  $l$ are set to unity while the acceleration of gravity is set to $g = 9.81$. The conjugate momentum is $p = m l \dot{\theta}$.
%Training data states are generated by numerical integration of \eqref{eq:singlePendulumDynamics} with a fourth-order RungeKutta scheme and a time-step of $0.1s$. 
The pendulum is simulated forward for 150s starting from four different initial angles, equally spaced between 0 and 180 degrees and with $0$ initial velocity. For our experiments we considered two training sets:  a full dataset  (f) that comprises a total of 600 data points, i.e. one data point per second, and  a  smaller dataset (s) with a total of only 64 data points, i.e. a data point every 10 seconds. For LNNs, the Lagrangian is expressed in terms of  the angular coordinates ($\theta$, $\dot{\theta})$. For HNNs, the  Hamiltonian  is expressed with respect to  $(\theta, p)$ and the CHNNs in terms of the  Cartesian  coordinates $(x,y,p_{x},p_{y})$. %\psk{i fail to see how the latter  is advantageous}. \\

The Hamiltonian/Lagrangian is parametrized with a MLP with two hidden layers, with 32 hidden units each and a softplus activation function. The network is trained on full batch data for 150000 epochs using the Adam \cite{Kingma.2014} optimization scheme. The learning rate is set to decay in two steps from $1e-2$ to $1e-3$ after the first $50000$ epochs and finally to $1e-4$ after another $50000$ epochs. 

For testing purposes,  we simulated the system  for $100s$ starting from ten random initial conditions that were \textbf{not contained} in the training data. As error metrics we chose the absolute value of the energy error $ \Delta E $, for which we report the mean value and the standard deviation, as well as  the maximum energy error $\max \Delta E$. The regularization parameter $\lambda_H$ was chosen by cross-validation to $\lambda_H=0.07$ for HNN and $\lambda_H=0.01$ for CHHN.\\
\begin{table}
\caption{Single pendulum error metrics for networks  trained on the full  dataset (f) and smaller dataset (s). All values are reported as percentages of the maximum potential energy.}
	\label{tab:singlePendulum_fullData_deltaH}
	\centering
	\begin{tabular}{lcccc}
		\toprule
		Scheme   & $ \Delta E_f$ & $\max \Delta E_f$ & $\Delta E_s$ & $\max \Delta E_s$ \\
		\midrule
		Baseline 		& 0.3706 $\pm$ 0.3767				  & 1.5115 & 2.4763 $\pm$ 2.1174				& 8.5754  \\%Baseline HNN
		HNN				& 0.0022 $\pm$ 0.0020				  & 0.0107 & 0.0934 $\pm$ 0.1208				& 0.6008 \\
		\textbf{HNN + H-Reg.}	& \textbf{0.0011 $\pm$ 0.0012} 				  & \textbf{0.0058} & \textbf{0.0181 $\pm$ 0.0261} 				& \textbf{0.1760}\\
		CHNN			& 0.0024 $\pm$ 0.0033				  & 0.0257 & 0.2920 $\pm$ 0.4904 				& 2.3802 \\
		\textbf{CHNN + H-Reg.}	& \textbf{0.0019 $\pm$ 0.0022}				  & \textbf{0.0108} & \textbf{0.0584 $\pm$ 0.1244}				& \textbf{1.1310}	\\
		LNN & 0.0043 $\pm$ 0.0039  & 0.0132 & 0.2091 $\pm$ 0.2307& 0.7806 \\
		\bottomrule
	\end{tabular}
\end{table}
The results are displayed in Table \ref{tab:singlePendulum_fullData_deltaH} where comparisons with LNNs are also reported. The physics-enhanced neural networks outperformed the baseline approach in all cases and the novel regularization term was able to significantly reduce both error metrics even further. The performance was superior for the HNN and the relative improvement was larger in the Small Data regime i.e. for the training data set (s).

\subsection{Double Pendulum}
The double pendulum is a chaotic nonlinear physical system which consists of two point mass pendulums that are attached to one another and subjected to a constant gravitational force. Each mass is attached to a rod and these are considered massless. The generalized coordinates describing the system's state are chosen to be the angles which the first and second rod form with the vertical direction. The geometrical parameters $l_1, l_2, m_1, m_2$ representing the rods' length and point masses are set to unity. 
The double pendulum's dynamics are described by a system of coupled differential equations. As for the single pendulum, a solution was obtained by numerical integration of
\begin{equation}
	\begin{aligned}
		\ddot{\theta}_1 + \frac{l_2}{l_1} \frac{m_2}{m_1 + m_2} \cos (\theta_1 - \theta_2) \ddot{\theta}_2 &= -\frac{l_2}{l_1} \frac{m_2}{m_1 + m_2} \dot{\theta}_2^2 \sin(\theta_1 - \theta_2) -  \frac{g}{l_1} \sin \theta_1 \\
		\ddot{\theta}_2 + \frac{l_1}{l_2} \cos(\theta_1 - \theta_2) \ddot{\theta}_1 &= \frac{l_1}{l_2} \dot{\theta}_1^2 \sin(\theta_1 - \theta_2) - \frac{g}{l_2}\sin \theta_2 \enspace.
	\end{aligned} 
	\label{eq:2pd}
\end{equation}

The MLP parametrization of the double pendulum's Hamiltonian/Lagrangian was chosen to consist of two hidden layers with 128 hidden units each and the softplus activation function. We trained the model on a full size dataset (f) with 6000 data points, i.e. ten data point per second, and small dataset (s) with 600 data points, i.e. one data point per second.  The network was trained on full batch data for 150000 epochs using Adam \cite{Kingma.2014} optimizer. The identical learning rate schedule as in the single pendulum case is used. During testing, we simulated the system for for $100s$ starting from  ten random initial conditions that were \textbf{not contained} in the training data. 
%A baseline black-box MLP representation of the system dynamics is again considered for comparison.
The regularization parameter $\lambda_H$ was chosen by cross-validation to $\lambda_H=0.2$ for HNN and $\lambda_H=0.005$ for CHHN.
\begin{table}
	\caption{Double pendulum  error metrics for networks  trained on the full  dataset (f) and the smaller dataset (s). All values are reported as percentages of the maximum potential energy. }
	\label{tab:doublePendulum_fullData_deltaH}
	\centering
	\begin{tabular}{lcccc}
		\toprule
		Scheme   & $ \Delta E_f$ & $\max \Delta E_f$ & $ \Delta E_s$ & $\max \Delta E_s$ \\
		\midrule
		Baseline 		& 14.201 $\pm$ 9.2108	& 48.267 & - & - 		\\%Baseline HNN
		HNN      		& 0.809 $\pm$	1.0147		& 4.962 & 18.393 $\pm$ 22.8308			& 186.648 \\
		\textbf{HNN + H-Reg.}    & \textbf{0.353 $\pm$ 0.4175} 			& \textbf{2.320} & \textbf{6.086 $\pm$ 6.6419} 			& \textbf{33.000} 		\\
		CHNN      		& 0.052	$\pm$ 0.0507 			& 0.250	 & 0.344 $\pm$ 0.4350 & 3.625 		\\
		\textbf{CHNN + H-Reg.}	& \textbf{0.021 $\pm$ 0.0227}	& \textbf{0.178}  & \textbf{0.166 $\pm$ 0.2004}	& \textbf{1.705}	\\
		LNN & 1.290 $\pm$ 2.2827  & 12.168 & - & -\\
		\bottomrule
	\end{tabular}
\end{table}

The results are displayed in Table \ref{tab:doublePendulum_fullData_deltaH} where comparisons with LNNs are also reported. The physics-enhanced neural networks outperformed the baseline approach \footnote{ For the small training set the baseline approach diverged} in all cases and the novel regularization parameter was able to significantly reduce both error metrics even further. This improvement was most striking in the Small Data regime. We note that the CHNN algorithm performed much better than HNN for this example and the LNN diverged for the small data case.

\section{Conclusions}
Augmenting the loss function of physics-enhanced neural networks with an additional regularization term can lead to better predictive capabilities, especially in the Small Data regime. We showed that this approach significantly outperforms the current methods for the single and double pendulum example and under initial conditions not contained in the training data.\\
A current disadvantage is that the regularization parameter $\lambda_{H}$ has to be chosen for each training data set separately. Here, further improvements based on advanced techniques regarding the training of PINNS \cite{Wang.2020,Wang.2020b} are possible.

\newpage
\printbibliography

@article{Battaglia.2018,
 author = {Battaglia, Peter W. and Hamrick, Jessica B. and Bapst, Victor and Sanchez-Gonzalez, Alvaro and Zambaldi, Vinicius and Malinowski, Mateusz and Tacchetti, Andrea and Raposo, David and Santoro, Adam and Faulkner, Ryan and others},
 year = {2018},
 title = {{Relational inductive biases, deep learning, and graph networks}},
 journal = {{arXiv preprint arXiv:1806.01261}}
}

@article{Cranmer.2020,
 author = {Cranmer, Miles and Greydanus, Sam and Hoyer, Stephan and Battaglia, Peter and Spergel, David and Ho, Shirley},
 year = {2020},
 title = {{Lagrangian neural networks}},
 journal = {{arXiv preprint arXiv:2003.04630}}
}

@inproceedings{Finzi.2020,
 author = {Finzi, Marc and Stanton, Samuel and Izmailov, Pavel and Wilson, Andrew Gordon},
 title = {{Generalizing convolutional neural networks for equivariance to lie groups on arbitrary continuous data}},
 pages = {3165--3176},
 booktitle = {{International Conference on Machine Learning}},
 year = {2020}
}

@article{Greydanus.2019,
 author = {Greydanus, Sam and Dzamba, Misko and Yosinski, Jason},
 year = {2019},
 title = {{Hamiltonian neural networks}},
 journal = {{arXiv preprint arXiv:1906.01563}}
}

@misc{JamesBradbury.2018,
 author = {{James Bradbury} and {Roy Frostig} and {Peter Hawkins} and {Matthew James Johnson} and {Chris Leary} and {Dougal Maclaurin} and {George Necula} and {Adam Paszke} and {Jake VanderPlas} and {Skye Wanderman-Milne} and {Qiao Zhang}},
 year = {2018},
 title = {{JAX: composable transformations of Python+NumPy programs}},
 url = {http://github.com/google/jax}
}

@article{Kingma.2014,
 author = {Kingma, Diederik P. and Ba, Jimmy},
 year = {2014},
 title = {{Adam: A method for stochastic optimization}},
 journal = {{arXiv preprint arXiv:1412.6980}}
}

@inproceedings{Maclaurin.2015,
 author = {Maclaurin, Dougal and Duvenaud, David and Adams, Ryan P.},
 title = {{Autograd: Effortless gradients in numpy}},
 pages = {5},
 volume = {238},
 booktitle = {{ICML 2015 AutoML Workshop}},
 year = {2015}
}

@misc{MartnAbadi.2015,
 author = {{Martin Abadi} and {Ashish Agarwal} and {Paul Barham} and {Eugene Brevdo} and {Zhifeng Chen} and {Craig Citro} and {Greg S. Corrado} and {Andy Davis} and {Jeffrey Dean} and {Matthieu Devin} and {Sanjay Ghemawat} and {Ian Goodfellow} and {Andrew Harp} and {Geoffrey Irving} and {Michael Isard} and {Yangqing Jia} and {Rafal Jozefowicz} and {Lukasz Kaiser} and {Manjunath Kudlur} and {Josh Levenberg} and {Dandelion Man{\'e}} and {Rajat Monga} and {Sherry Moore} and {Derek Murray} and {Chris Olah} and {Mike Schuster} and {Jonathon Shlens} and {Benoit Steiner} and {Ilya Sutskever} and {Kunal Talwar} and {Paul Tucker} and {Vincent Vanhoucke} and {Vijay Vasudevan} and {Fernanda Vi{\'e}gas} and {Oriol Vinyals} and {Pete Warden} and {Martin Wattenberg} and {Martin Wicke} and {Yuan Yu} and {Xiaoqiang Zheng}},
 year = {2015},
 title = {{TensorFlow: Large-scale machine learning on heterogeneous systems}},
 url = {https://www.tensorflow.org/}
}

@article{SanchezGonzalez.2019,
 author = {Sanchez-Gonzalez, Alvaro and Bapst, Victor and Cranmer, Kyle and Battaglia, Peter},
 year = {2019},
 title = {{Hamiltonian graph networks with ode integrators}},
 journal = {{arXiv preprint arXiv:1909.12790}}
}

@article{Scarselli.2008,
 author = {Scarselli, Franco and Gori, Marco and Tsoi, Ah Chung and Hagenbuchner, Markus and Monfardini, Gabriele},
 year = {2008},
 title = {{The graph neural network model}},
 pages = {61--80},
 volume = {20},
 number = {1},
 journal = {{IEEE Transactions on Neural Networks}}
}

@article{Wang.2020,
 author = {Wang, Sifan and Teng, Yujun and Perdikaris, Paris},
 year = {2020},
 title = {{Understanding and mitigating gradient pathologies in physics-informed neural networks}},
 journal = {{arXiv preprint arXiv:2001.04536}}
}

@article{Wang.2020b,
 author = {Wang, Sifan and Yu, Xinling and Perdikaris, Paris},
 year = {2020},
 title = {{When and why PINNs fail to train: A neural tangent kernel perspective}},
 journal = {{arXiv preprint arXiv:2007.14527}}
}

@article{lusch2018deep,
  title={Deep learning for universal linear embeddings of nonlinear dynamics},
  author={Lusch, Bethany and Kutz, J Nathan and Brunton, Steven L},
  journal={Nature communications},
  volume={9},
  number={1},
  pages={1--10},
  year={2018},
  publisher={Nature Publishing Group}
}

@article{zhu2019physics,
  title={Physics-constrained deep learning for high-dimensional surrogate modeling and uncertainty quantification without labeled data},
  author={Zhu, Yinhao and Zabaras, Nicholas and Koutsourelakis, Phaedon-Stelios and Perdikaris, Paris},
  journal={Journal of Computational Physics},
  volume={394},
  pages={56--81},
  year={2019},
  publisher={Elsevier}
}

@article{raissi2019physics,
  title={Physics-informed neural networks: A deep learning framework for solving forward and inverse problems involving nonlinear partial differential equations},
  author={Raissi, Maziar and Perdikaris, Paris and Karniadakis, George E},
  journal={Journal of Computational Physics},
  volume={378},
  pages={686--707},
  year={2019},
  publisher={Elsevier}
}

@article{kaltenbach2020incorporating,
  title={Incorporating physical constraints in a deep probabilistic machine learning framework for coarse-graining dynamical systems},
  author={Kaltenbach, Sebastian and Koutsourelakis, Phaedon-Stelios},
  journal={Journal of Computational Physics},
  volume={419},
  year={2020},
  publisher={Elsevier}
}

@article{lutter2019deep,
  title={Deep lagrangian networks: Using physics as model prior for deep learning},
  author={Lutter, Michael and Ritter, Christian and Peters, Jan},
  journal={arXiv preprint arXiv:1907.04490},
  year={2019}
}

@article{toth2019hamiltonian,
  title={Hamiltonian generative networks},
  author={Toth, Peter and Rezende, Danilo Jimenez and Jaegle, Andrew and Racani{\`e}re, S{\'e}bastien and Botev, Aleksandar and Higgins, Irina},
  journal={arXiv preprint arXiv:1909.13789},
  year={2019}
}

@article{kaltenbach_2021,
  author =        {Sebastian Kaltenbach and Phaedon-Stelios Koutsourelakis},
  journal =       {ICLR},
  title =         {Physics-aware, probabilistic model order reduction with guaranteed stability},
  year =          {2021},
}

@article{walters_2021,
  author =        {Robin Walters and  Jinxi Li and Rose Yu},
  journal =       {ICLR},
  title =         {Trajectory prediction using equivariant continuous convolution},
  year =          {2021},
}

\end{document}